\documentclass[runningheads]{llncs}
\usepackage[T1]{fontenc}
\usepackage{graphicx}
\usepackage{booktabs}
\usepackage[misc]{ifsym}

\begin{document}

\title{Advancing Solar Flare Prediction using Deep Learning with Active Region Patches}

\titlerunning{Advancing Solar Flare Prediction}

\author{Chetraj Pandey\Letter \orcidID{0000-0002-4699-4050} \and Temitope Adeyeha \orcidID{0000-0001-9512-8733}  \and Jinsu Hong \orcidID{0009-0002-4383-1376}  \and Rafal A. Angryk\orcidID{0000-0001-9598-8207} \and 
Berkay Aydin\orcidID{0000-0002-9799-9265}}
\authorrunning{C. Pandey et al.}
\institute{Georgia State University, Atlanta, GA, 30303, USA\\
\email{\{cpandey1, tadeyeha1, jhong36, rangryk, baydin2\} @gsu.edu}}
\toctitle{Advancing Solar Flare Prediction using Deep Learning with Active Region Patches}
\tocauthor{Chetraj~Pandey, Temitope~Adeyeha, Jinsu~Hong, Rafal A.~Angryk, Berkay~Aydin}

\maketitle              % typeset the header of the contribution

\begin{abstract}
Solar flares are one of the key space weather phenomena characterized by sudden and intense emissions of radiation from the Sun. The precise and reliable prediction of these phenomena is important due to their potential adverse effects on both space and Earth-based infrastructure. In this paper, we introduce a novel methodology for leveraging shape-based characteristics of magnetograms of active region (AR) patches and provide a novel capability for predicting solar flares covering the entirety of the solar disk (AR patches spanning from -90$^{\circ}$ to +90$^{\circ}$ of solar longitude). We create three deep learning models: (i) ResNet34, (ii) MobileNet, and (iii) MobileViT to predict $\geq$M-class flares and assess the efficacy of these models across various ranges of solar longitude. Given the inherent imbalance in our data, we employ augmentation techniques alongside undersampling during the model training phase, while maintaining imbalanced partitions in the testing data for realistic evaluation. We use a composite skill score (CSS) as our evaluation metric,  computed as the geometric mean of the True Skill Score (TSS) and the Heidke Skill Score (HSS) to rank and compare models. The primary contributions of this work are as follows: (i) We introduce a novel capability in solar flare prediction that allows predicting flares for each ARs throughout the solar disk and evaluate and compare the performance, (ii) Our candidate model (MobileNet) achieves a CSS=0.51 (TSS=0.60 and HSS=0.44), CSS=0.51 (TSS=0.59 and HSS=0.44), and CSS=0.48 (TSS=0.56 and HSS=0.40) for AR patches within ±30$^{\circ}$, ±60$^{\circ}$, ±90$^{\circ}$ of solar longitude respectively. Additionally, we demonstrate the ability to issue flare forecasts for ARs in near-limb regions (regions between ±60$^{\circ}$ to ±90 $^{\circ}$) with a CSS=0.39 (TSS=0.48 and HSS=0.32), expanding the scope of AR-based models for solar flare prediction. This advancement opens new avenues for more reliable prediction of solar flares, thereby contributing to improved forecasting capabilities.

\end{abstract}

\section{Introduction}

Solar flares are temporary events characterized by abrupt and massive eruptions of radiation from the Sun's surface. They are critical space weather phenomena with significant implications for both space-based and Earth-based infrastructures. The National Oceanic and Atmospheric Administration (NOAA) classifies solar flares into five classes based on their peak X-ray flux levels: A, B, C, M, and X, representing the flares from weakest to strongest \cite{spaceweather}. Flares weaker than A-class are typically not detected and are thus considered flare-quiet (FQ). M- and X-class flares are the strongest and can cause near-Earth impacts, including disruptions in electrical power grids, the aviation industry, radio and satellite communications, and pose radiation hazards to astronauts in space. Therefore, developing precise and reliable methods to predict solar flares is necessary to mitigate the potential adverse effects of space weather on Earth.

Active regions (ARs) are the areas of high activity on the Sun's surface, noted for their intense magnetic fields concentrated within sunspots. These magnetic fields often undergo significant distortion and instability, triggering plasma disturbances and releasing energy in the form of flares and other solar phenomena \cite{Toriumi2019}. This makes ARs the regions of interest, emphasizing the importance of utilizing AR-based features for predicting solar flares. However, the magnetic field measurements, which are the dominant feature employed by AR-based methods, are susceptible to severe projection effects caused by the orientation of the observing instrument relative to the solar surface. Therefore, as ARs approach the solar limbs, specifically beyond $\pm$60$^{\circ}$ of solar longitude, the magnetic field readings become distorted \cite{Falconer2016}, which limits the existing models to include data pertaining to central locations only \cite{pandey2021bigdata}, \cite{Pandey2022}, \cite{pandeyecml2023}. To address this, we derive images from original line-of-sight (LoS) magnetogram rasters of AR patches with our novel data processing pipeline that captures the overall morphology and spatial distribution of active regions, retaining important shape-based parameters such as size, directionality, sunspot borders, and polarity inversion lines \cite{ji2023systematic}, \cite{mcintosh1990classification}.  We recognize the persistence of  projection effects also in images of AR magnetogram patches; however, we hypothesize that the complex feature learning capabilities of contemporary deep learning models can potentially learn from the shape-based features retained in images while filtering the distorted readings. Consequently, we include data from ARs beyond $\pm$60$^{\circ}$ as well, thereby providing a novel capability to predict solar flares across the entire disk.

Furthermore, it is important to note that the tracked AR patches vary in size depending on the size of the ARs. Existing approaches have been limited to AR patches in central locations, often resizing rectangular patches to obtain square images. However, this resizing distorts the original aspect ratio, consequently altering the shapes and sizes of ARs. Alternatively, variable-sized AR patches are cropped (using methods like center crop or random crop) to obtain square images, resulting in information loss. In contrast, we propose and utilize a sliding window kernel-based approach. This method select such a cropped region that maximizes total unsigned flux (USFLUX: the sum of the absolute of the magnetic field strength values), maintaining the original aspect ratios of AR patches and preserving critical spatial features. By maximizing the USFLUX, we ensure that we extract the most representative region with significant magnetic flux build up. This method adapts to the variability in AR patch shapes and sizes, avoiding distortion and prioritizing the capture of more relevant information. 

In this study, we develop deep learning models to predict solar flares of magnitude $\geq$M-class by leveraging images created from cutouts (patches) of magnetograms corresponding to ARs. We employ these images to train three distinct deep learning architectures: ResNet34 \cite{resnet}, MobileNet \cite{mobilenet-v3}, and MobileViT \cite{MobileViTLG}. Our contributions can be summarized as follows: (i) we propose a novel methodology for preprocessing magnetograms of AR patches, focusing on preserving shape-based parameters while maximizing the total unsigned flux (USFLUX) \cite{Bobra2014} at the instance level, (ii) we showcase that our models are capable of predicting flares across the entire solar disk, including often overlooked near-limb regions, improving the comprehensiveness of AR-based solar flare prediction models, and (iii) through rigorous experimentation, our models demonstrate superior performance compared to existing approaches, significantly contributing to ongoing efforts aimed at enhancing space weather forecasting capabilities.

The remainder of this paper is organized as follows: In Sec.~\ref{sec:rel}, we outline the various approaches used in solar flare prediction along with contemporary work using deep learning methods. In Sec.~\ref{sec:datamodel}, we explain our data processing pipeline, class-wise distribution of the data for binary prediction mode along with description of our flare prediction models. In Sec.~\ref{sec:expt}, we present our experimental design, evaluations, and discuss the implications of our work, lastly, in Sec.~\ref{sec:conc}, we provide our concluding remarks and discuss avenues for future work.

\section{Related Work}\label{sec:rel}

A range of methodologies, such as human-based prediction techniques (e.g., \cite{Crown2012}), statistical approaches (e.g., \cite{Lee2012}), and numerical simulations based on physics-based models (e.g., \cite{Kusano2020}), have been employed to predict solar flares. Recently, the success of data-driven approaches, which leverage machine learning and deep learning techniques, has significantly increased owing to their capacity to exploit extensive datasets \cite{Hong2023CogMi} and their experimental achievements \cite{Nishizuka2018}. As solar flares are phenomena caused by sudden, abrupt changes in the magnetic field in the solar atmosphere, these data-driven approaches most commonly utilizes magnetogram-based data  which includes solar full-disk magnetograms (e.g., \cite{HongICMLA2023}, \cite{PandeyICMLA2023}, \cite{Pandey2023DS}, \cite{PandeyAIKE2023}, \cite{Pandey2023DSAA}), multivariate time series (MVTS) data extracted from solar photospheric vector magnetograms (e.g., \cite{Ji2023}, \cite{Ji2022}), cutouts or patches of tracked AR (e.g., \cite{Huang2018}, \cite{Li2020}), and features summarizing each AR patches (e.g., \cite{Bloomfield2012}, \cite{Bobra2015}).

A deep learning model based on a multi-layer perceptron to predict solar flares $\geq$C and $\geq$M class  was presented in \cite{Nishizuka2018}. In this study, they used 79 manually selected features extracted from multi-modal solar observations of full solar disk, which included vector magnetograms and extreme ultraviolet (EUV) images to predict $\geq$M- and $\geq$C-class flares.  In \cite{Park2018}, a CNN-based hybrid model to predict the occurrence of a $\geq$C-class flares. Similarly, in \cite{pandeyecml2023}, \cite{Pandey2023DS} we presented a convolutional neural network (CNN) based model to predict $\geq$M-class flares utilizing full-disk magnetogram images. While these full-disk models includes near-limb regions, by themselves they are unable to localize the relevant AR which is likely to flare and instead issue one single forecast for entire solar disk. 

In \cite{Bobra2015}, a support vector machine (SVM) based model trained with 25 AR summary parameters extracted from vector magnetograms of AR patches within $\pm$68$^\circ$ of solar longitude (central meridian distance) was presented. Similarly, in \cite{Ji2022}, a deep learning based time series classifier and in \cite{Ji2023} a sliding window Time Series Forest (TSF) was trained with a MVTS data of 24 space weather related physical parameters primarily calculated from AR magnetograms with in $\pm$70$^\circ$ of solar longitude. Furthermore, a CNN-based flare forecasting model trained with AR patches (resized to 100$\times$100 pixels) extracted from LoS magnetograms within $\pm$30$^{\circ}$ of the solar longitude to predict $\geq$C-, $\geq$M-, and $\geq$X-class flares was presented in \cite{Huang2018}. More recently, \cite{Li2023} proposed a CNN-based model named ``CARFFM-4'' trained with AR patches (sized to 160$\times$160 pixels) created from R parameter \cite{Schrijver2007} within $\pm$30$^{\circ}$ of solar longitude to predict $\geq$M-class flares in next 48 hours. It is important to note that, there is variability in literature in terms of type of data modality which includes multiple instruments (HMI/SDO, AIA/SDO, MDI/SOHO) and data types (EUV images, magnetograms and extracted features corresponding to AR and full-disk). Furthermore the variability in prediction targets ($\geq$C-, $\geq$M-, $\geq$X-class flares) and forecasting horizon (24 hours and 48 hours) is also prominent. The predictive capabilities of AR-based models are often limited by observations taken from central locations from $\pm30^{\circ}$ to $\pm70^{\circ}$. The full-disk models complement the issue of longitudinal coverage in AR-based models; however, they fail to pin-point an active region and issue a single forecast for the entire solar disk. In this work, we introduce a limb-to-limb AR-based flare prediction model encompassing full 180$^{\circ}$ ($\pm90^{\circ}$ of the solar longitude) and evaluate our models efficacy in different longitudinal range and provide a novel capability, to our knowledge, missing in operational systems.  
 
\section{Data and Model}\label{sec:datamodel}

\begin{figure}[tbh!]
\centering
\includegraphics[width=0.95\linewidth ]{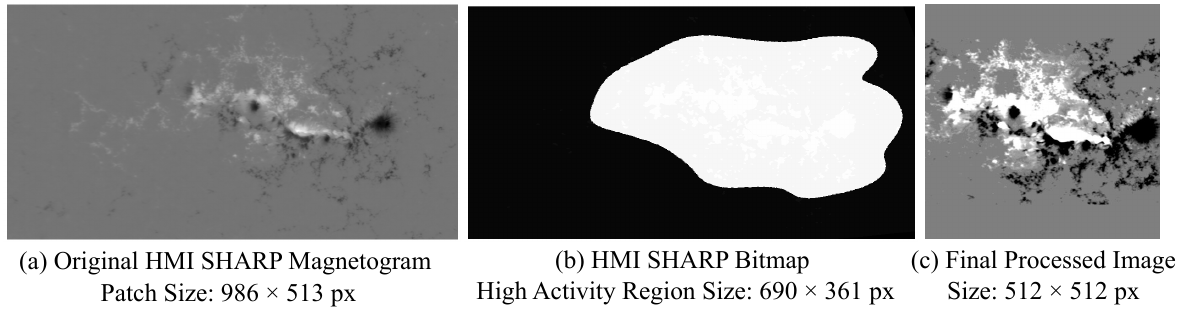}
\caption[]{An illustrative example of (a) Original raw input magnetogram of HMI AR patch corresponding to HARP number: 4781 (NOAA AR number: 12205) at timestamp 2014-11-06T18:00:00 UTC, (b) Bitmap corresponding to HMI AR patch in (a) showing the region of interest indicated by white pixels, (c) Final processed image of AR patch in (a) now sized to 512$\times$512, that is used to train our models.}
\label{fig:dataexample}

\end{figure}

The raw input data in our work includes LoS magnetograms of ARs provided by the Helioseismic and Magnetic Imager (HMI) \cite{Schou2011} onboard the Solar Dynamics Observatory (SDO) \cite{Pesnell2011}, which are publicly available as a data product named Space-Weather HMI Active Region Patches (SHARP) \cite{Bobra2014} from the Joint Science Operations Center\footnote{http://jsoc.stanford.edu}. In this work, we utilized magnetograms spanning from May 2010 to 2018, sampling magnetograms at a cadence of one hour. The magnetograms of AR patches contain rasters of magnetic field strength values typically ranging from $\sim$$\pm$4500 G. An example of magnetogram of AR patch is shown in Fig.~\ref{fig:dataexample}(a). Along with magnetograms, we use bitmaps (another data product from the SHARP series) which define the region with pixels located within or outside the ARs, providing the region of interest within the AR patch as shown in Fig.~\ref{fig:dataexample} (b). The bitmap are equal in size to the LoS magnetograms of AR patches and contains five unique pixel values: {0, 1, 2, 33, 34}, where the values 33 and 34 indicate the pixels that are within the AR region and hence our region of interest \cite{Bobra2014}. For each AR patch, we assign a binary label using peak X-ray flux converted to NOAA flare classes such that: (i) $\geq$M indicates Flare (FL) signifying the existence of a relatively strong flaring activity, and (ii) $<$M indicates No Flare (NF) with a prediction window of 24 hours. To elaborate, from the timestamp of an AR patch to the next 24 hours, if the maximum flare class is $<$M, then we label the AR patch as NF; otherwise, FL.

\begin{figure}[tbh!]
\centering
\includegraphics[width=0.95\linewidth ]{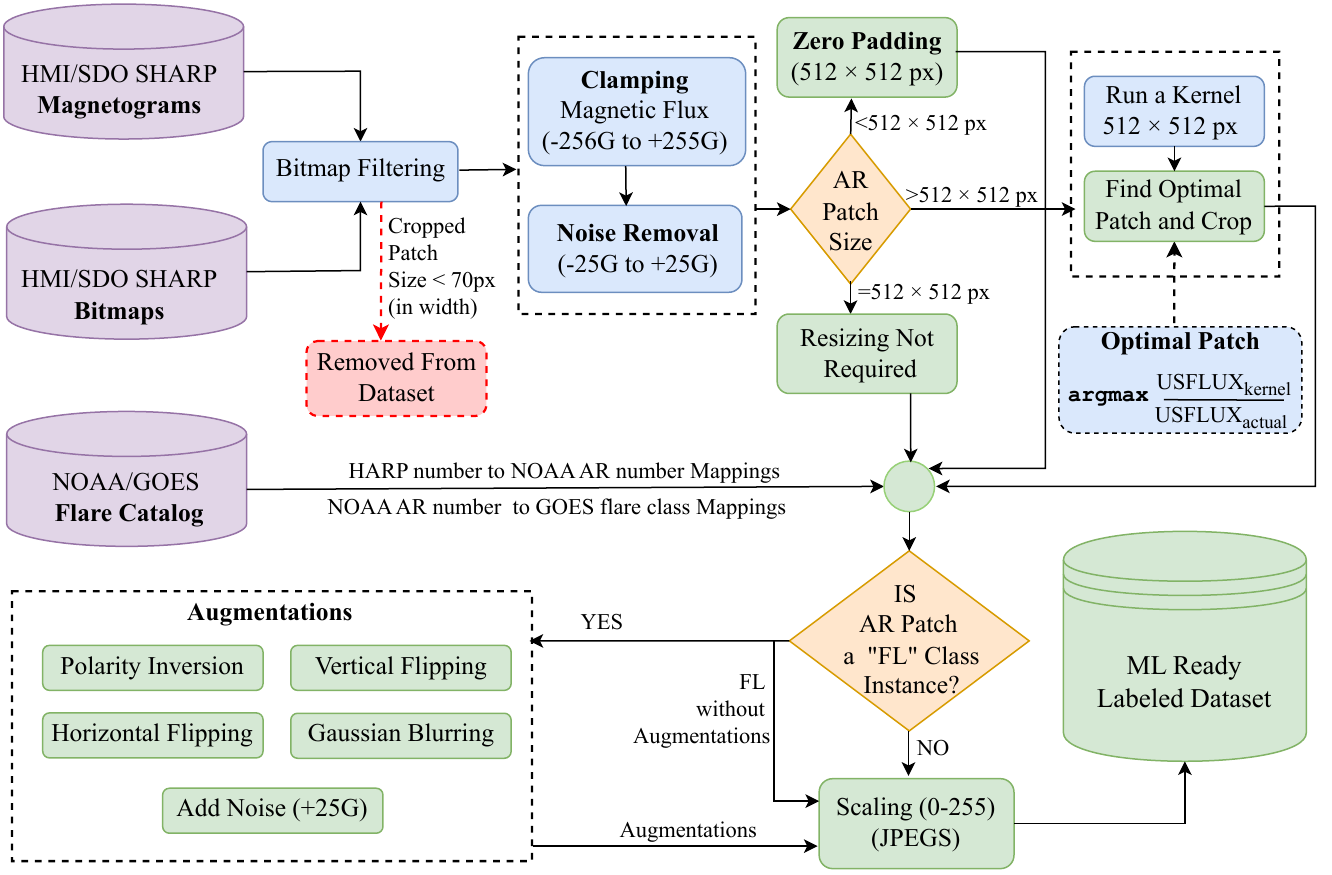}
\caption[]{The process flow diagram of data processing pipeline used in this work. It shows a sequential pipeline for creating JPEG images from magnetogram rasters and corresponding bitmaps along with data augmentation pipeline given the label for the magnetogram patch. Boxes colored in green collectively defines our entire dataset.}
\label{fig:datapipeline}

\end{figure}

In our data processing pipeline, which is illustrated in Fig.~\ref{fig:datapipeline}, we begin by collecting hourly instances of original raw input magnetograms of active region (AR) patches, alongside their corresponding bitmaps. Our initial step involves applying the bitmap as a filter to precisely crop the AR patches, isolating the regions with high activity. Subsequently, we implement a size filter: if the resulting cropped AR patches are smaller than 70 pixels in width, we exclude them from our dataset. It is worth noting that we determine this threshold based on the overall data distribution, ensuring retention of all instances corresponding to `FL'  instances while removing those from the `NF'  class. Following this filtering stage, we proceed to adjust the magnetic flux. We cap the flux values at $\pm$256G, and any flux values within $\pm$25G are set to 0 to mitigate noise. Ensuring uniformity in size, we apply zero-padding to patches smaller than 512$\times$512 pixels. Conversely, for larger patches, exceeding 512$\times$512 pixels, we employ a 512$\times$512 kernel to select the patch with the maximum total unsigned flux (namely USFLUX, which is the sum of the absolute value of magnetic field strength represented as raster values in magnetograms). By doing this, we aim to minimize information loss by picking a spatial window where the total flux is the highest, which is more likely to include the regions of interest. Finally, to standardize the representation, all patches are scaled to fit within the range of 0-255, facilitating the generation of images. An example of final preprocessed image utilizing the  magnetogram raster in Fig.~\ref{fig:dataexample} (a) and bitmap in Fig.~\ref{fig:dataexample} (b) is  shown in Fig.~\ref{fig:dataexample} (c).

\begin{figure*}[tbh!]
\begin{tabular}{c c}
\includegraphics[width=0.48\linewidth]{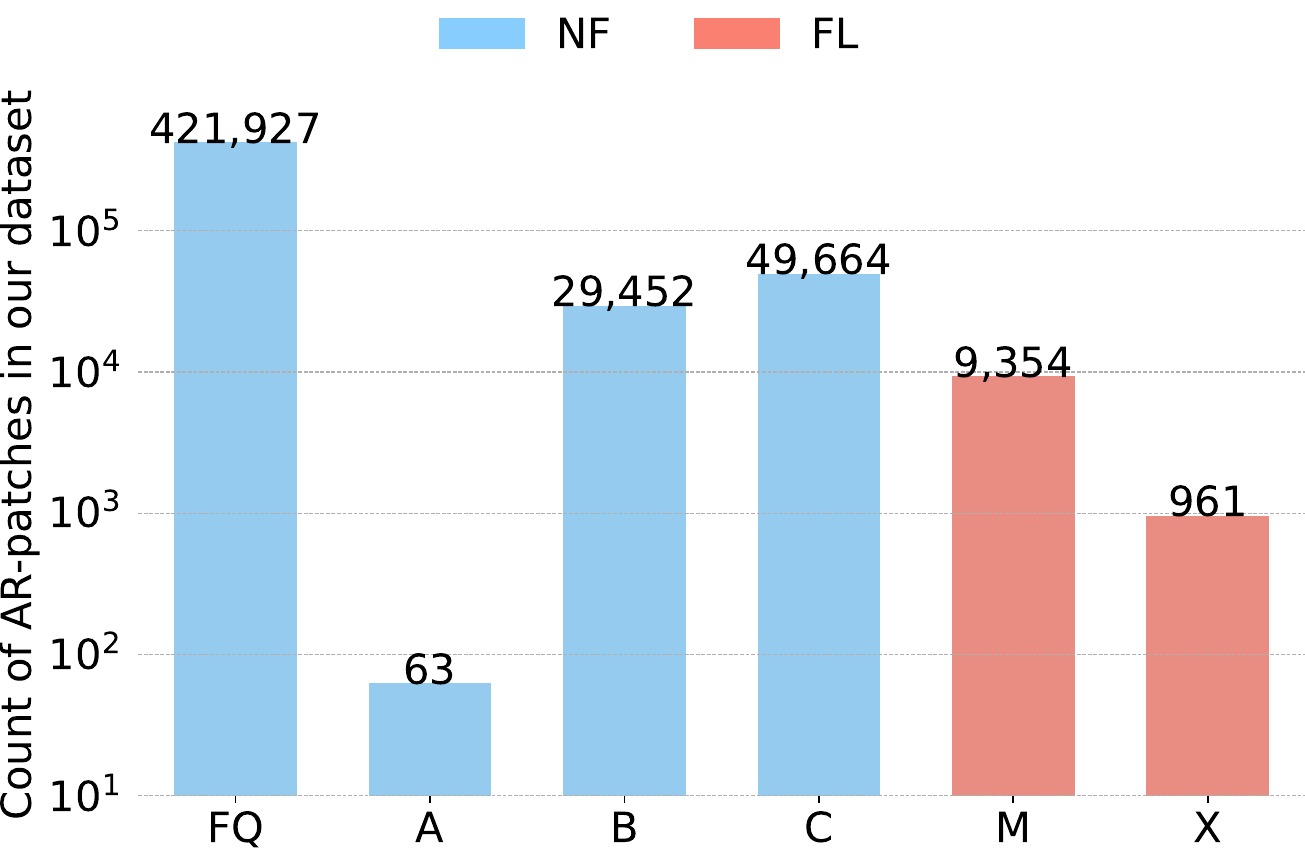}&
\includegraphics[width=0.48\linewidth]{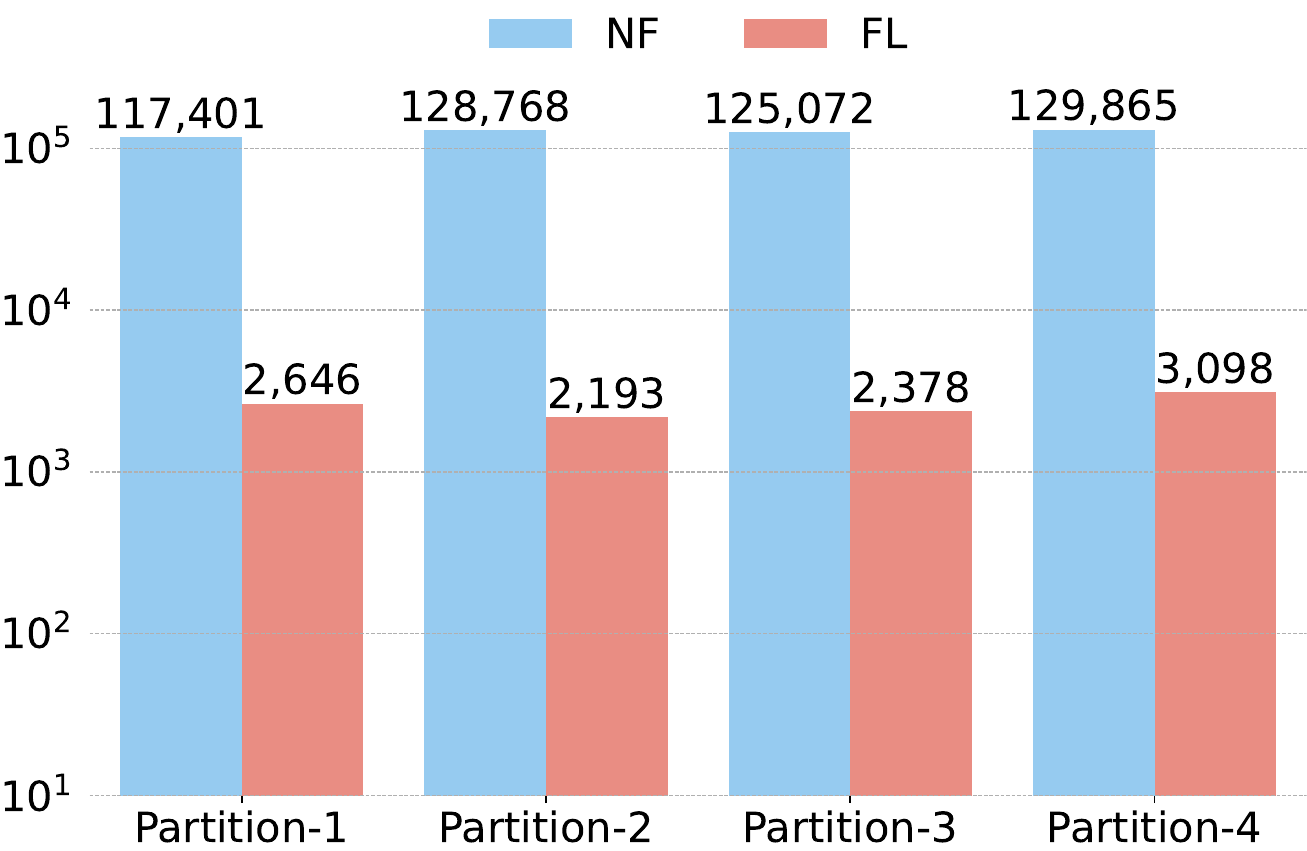}\\
(a) & (b)
    \end{tabular}
\caption{(a)Overall data distribution (b) Overall data partitioned into four tri-monthly partitions. Note: The height of the bars are in log scale.}
\label{fig:data}

\end{figure*}

The overall distribution of our binary-labeled AR patches data, with flare classes NF (comprising flare-quiet (FQ), A-, B-, and C-class flares) and FL (including M- and X-class flares), is depicted in Fig.~\ref{fig:data}(a). In total, we have 501,106 instances belonging to the NF class and 10,315 instances belonging to the FL class, resulting in a class imbalance ratio of $\sim$ 1:49. We split our entire dataset using temporally non-overlapping tri-monthly partitioning into four partitions based on the onset timestamp of HARP series. To elaborate, we ensure that the data corresponding to an entire AR trajectory is included in only one partition to prevent double-dipping by using the onset timestamp of the HARP series for tri-monthly partitioning, in contrast to using the observation timestamp of magnetograms as mentioned in \cite{pandey2021bigdata}. The resulting data distribution is shown in Fig.~\ref{fig:data} (b), where we use Partitions 1 and 2 as our training set while Partitions 3 and 4 as validation and test set respectively. The data augmentation and undersampling of training set is later described in Sec.~\ref{sec:exptsetup}.

The task of solar flare prediction in this work is formalized as a binary image classification problem; therefore, we select three general-purpose deep learning models: (i) ResNet34 \cite{resnet}, (ii) MobileNet \cite{mobilenet-v3}, and (iii) MobileViT \cite{MobileViTLG}. Recently, attention-based models, notably Vision Transformers (ViTs) \cite{vit}, have emerged as frontrunners in the task of image classification. They have showcased superior performance on large-scale datasets compared to standard CNNs. However, ViTs typically boast a high number of trainable parameters (ranging from approximately 86 to 632 million), making them demanding in terms of computational resources. This limits their practicality in scenarios with restricted computational capabilities or smaller datasets. In response to these challenges, particularly in our context with a modest dataset, we explore alternative models that strike a balance between accuracy and efficiency. Our focus lies in selecting lightweight architectures that can deliver competitive performance while being more resource-efficient. ResNet34 \cite{resnet} and MobileNet \cite{mobilenet-v3} are standard CNNs with $\sim$21.2 and $\sim$4.2 million trainable parameters, respectively, while MobileViT boasts $\sim$2 million trainable parameters. ResNet34 excels in capturing fine-grained details within images, potentially aiding in identifying subtle patterns indicative of solar flares. MobileNet, renowned for its efficiency, offers a balance between computational resources and accuracy, making it suitable for deployment in resource-limited environments such as space-based solar observation platforms. Additionally, MobileViT \cite{MobileViTLG}, a variant of the Vision Transformer \cite{vit} inspired by MobileNet is optimized for computational efficiency, leveraging both efficiency and global context capturing capabilities, potentially enhancing the models' performance in capturing complex patterns relevant to solar flare prediction. This selection of models aims to evaluate their efficacy for solar flare prediction while prioritizing lightweight architectures with varying designs.

\section{Experimental Evaluation} \label{sec:expt}

\subsection{Experimental Settings}\label{sec:exptsetup}
\begin{figure}[tbh!]
\centering
\includegraphics[width=0.95\linewidth ]{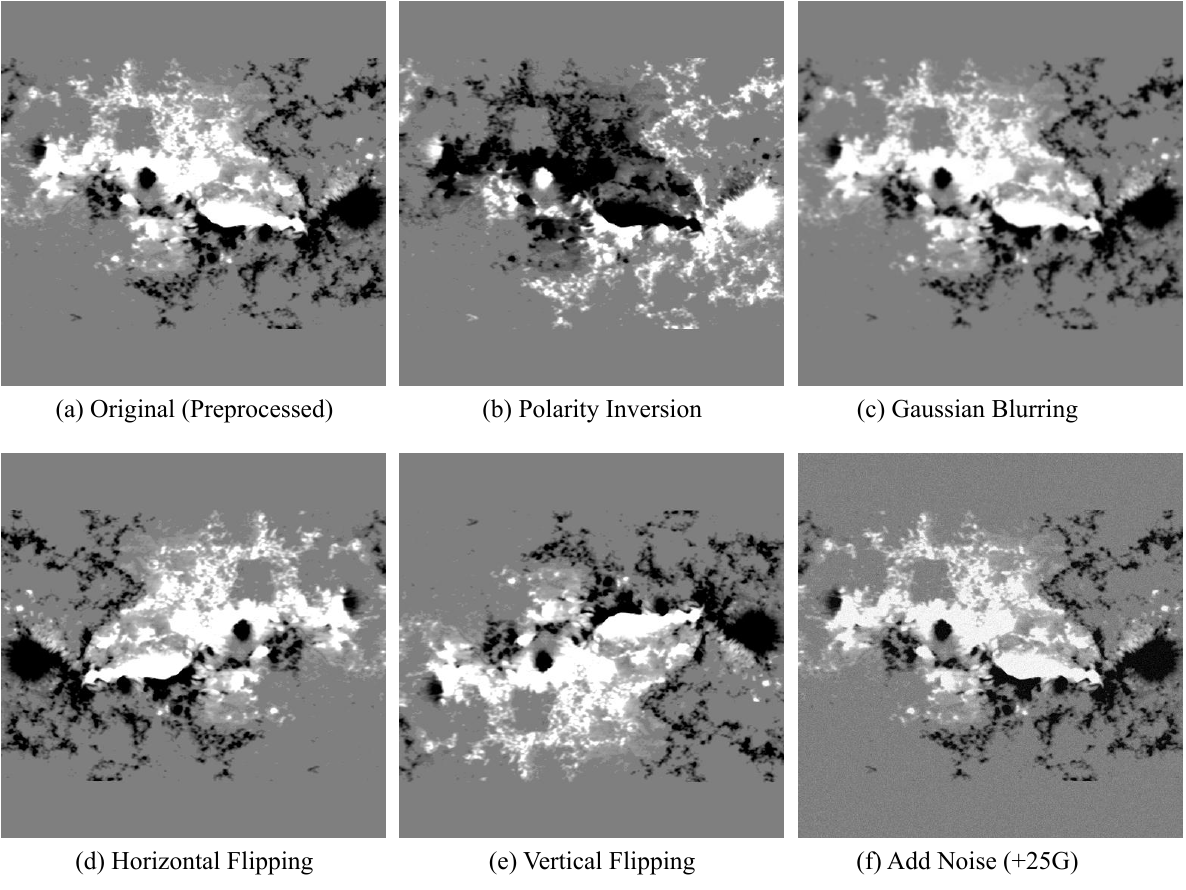}
\caption[]{An illustrative example of (a) input magnetogram of HMI AR patch corresponding to HARP number: 4781 (NOAA AR number: 12205) at timestamp 2014-11-06T18:00:00 UTC, (b-f) five different augmentations applied to AR patch in (a).}
\label{fig:aug}
% \vspace{-15pt}
\end{figure}

\begin{figure}[tbh!]
\centering
\includegraphics[width=0.95\linewidth ]{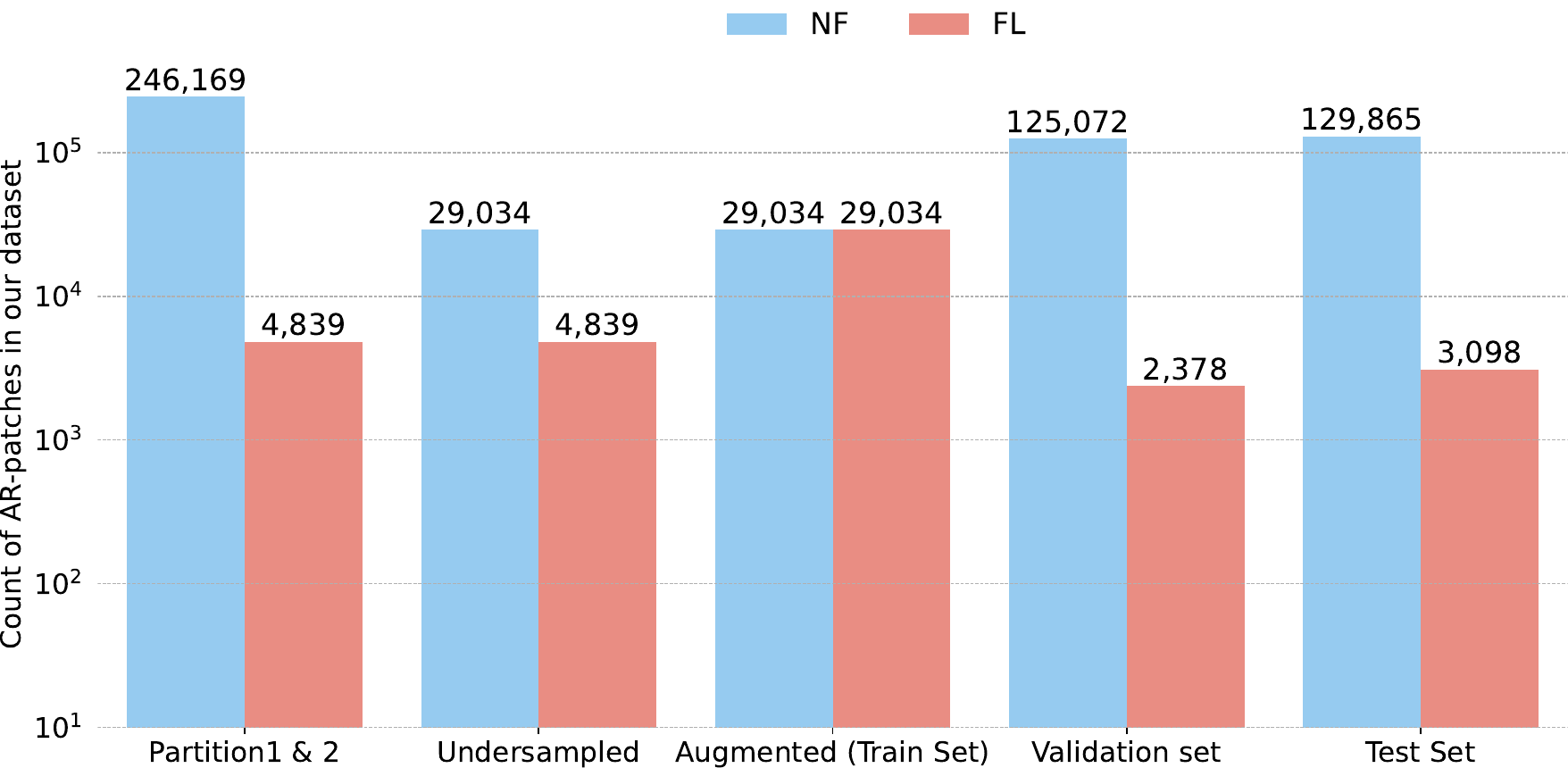}
\caption[]{Overall data partitioned into train set (showing original, undersampled and augmented data counts) , validation set, and test set used in this study.}
\label{fig:dataset}
% \vspace{-10pt}
\end{figure}

As mentioned earlier in Sec. \ref{sec:datamodel}, we follow time-segmented tri-monthly partitioning scheme to create four partitions of our entire dataset. Partition-1 and 2 combined are used as the training set. However, due to significant class imbalance in our dataset, we used undersampling together with data augmentation to create a balanced training set. Firstly, we augmented data instances belonging to the FL-class in our training set using five data augmentation techniques: (i) polarity inversion, which swaps the signs of positive polarity to negative and vice versa as shown in Fig. \ref{fig:aug} (b), (ii) Gaussian filtering, which applies a Gaussian blur to the image to reduce noise and detail (Fig. \ref{fig:aug} (c)), (iii) Horizontal Flipping, which involves flipping the image along a vertical axis (Fig. \ref{fig:aug} (d)), (iv) Vertical Flipping , which involves flipping the image along a horizontal axis (Fig. \ref{fig:aug} (e)), and (v) Adding random noise within $\pm$25G (Fig. \ref{fig:aug} (f)). To balance the FL-Class instances with NF, we undersampled our training data by randomly selecting 30\% of instances belonging to A-, B-, C-class flares each, and $\sim$8\% of instances from FQ from both Partition-1 and 2. For realistic evaluation, we maintained the original imbalanced distribution in Partitions 3 and 4, which are our validation and test sets respectively, as shown in Fig.~\ref{fig:dataset}.\\

\noindent \textbf{Model Parameters: } In our model hyperparameter selection process, we define the hyperparameter space, encompassing initial learning rate sets, weight decay sets, batch size sets, and class weight sets as shown in Table~\ref{table:params}. We also investigate three sets of class weights, although the training data was balanced, aiming to assess their impact on minimizing false positives. These class weight configurations are indicated as NF:FL, where 1:1 would suggest equal weights in loss minimization. To prioritize the minimization of false positives, we explored increasing the weights for NF-class instances. Following the definition of our hyperparameter space, we conduct a meticulous grid search across this space, evaluating on the validation set for all three models. During this search, we train our models using stochastic gradient descent (SGD) with cross-entropy loss. Additionally, we employ a  dynamic learning rate strategy, which reduced the initial learning rate by a factor of 10 every 10 epochs. Upon completing the grid search and evaluating the models, we identified the optimal hyperparameters as shown in Table~\ref{table:params}. These parameters exhibited superior performance, we use these to train our final models for 50 epochs and evaluate on the test set. 

\begin{table}[tbh]
\setlength{\tabcolsep}{6pt}
\renewcommand{\arraystretch}{1.2}
% \vspace{-15pt}
\caption{Hyperparameters search space with optimal hyperparamters for each model.}
\begin{center}
 \begin{tabular}{r r r r r}
\hline
\multicolumn{2}{r}{}              %% <-- mistake
&                                            %% <--  addition
\multicolumn{3}{c}{Optimal Parameters}\\
          %% <--  Changed
Hyperparameters & Search Space & ResNet34  & MobileNet  & MobileViT \\
\hline
%  &    &     &    & &    & \\
Learning Rate &  \{0.00001 to 0.01\}   &  0.001  & 0.001  & 0.001 \\

Weight Decay &  \{0.00001 to 0.01\}  &  0.01 & 0.01  & 0.001 \\

Batch Size & \{48, 64\} & 48 & 48 & 48 \\ 

Class Weights& \{1:1, 3:1, 5:1\} & 5:1 & 5:1 & 5:1 \\ 
\hline
\end{tabular}
% \vspace{}
\end{center}
\label{table:params}
% \vspace{-2em}
\end{table}

\noindent \textbf{Evaluation Metrics: } True Skill Statistic (TSS, in Eq.~\ref{eq:TSS}) and Heidke Skill Score (HSS, in Eq.~\ref{eq:HSS}), derived from the four elements of confusion matrix: TP, TN, FP, FN are the two forecast skills scores widely used in evaluating flare prediction models. 

\begin{equation}\label{eq:TSS}
    TSS = \frac{TP}{TP+FN} - \frac{FP}{FP+TN} 
\end{equation}
\begin{equation}\label{eq:HSS}
    HSS = 2\times \frac{TP \times TN - FN \times FP}{((P \times (FN + TN) + (TP + FP) \times N))}
\end{equation}
\begin{center}
    % \vspace{2pt}
where, $N = TN + FP$ and  $P = TP + FN$.    
\end{center}

TSS and HSS values range from -1 to 1, where 1 indicates all correct predictions, -1 represents all incorrect predictions (also, it means that all inverse predictions are correct, i.e., there is a skill), and 0 represents no skill. In contrast to TSS, HSS is an imbalance-aware metric that is commonly employed in solar flare prediction models due to the prevalent high class-imbalance ratios. However, choosing a candidate model based on two skill scores becomes difficult, as it demands preference of one metric over another at the end. Therefore, by combining TSS and HSS in a geometric mean as in the Composite Skill Score (CSS, in Eq.~\ref{eq:CSS}), we obtain a single metric that balances between discrimination ability and imbalance awareness. 

\begin{equation}\label{eq:CSS}
CSS = \left\{
\begin{array}{ll}
    0 & \quad \mbox{if   } TSS \times HSS < 0 \\
    \sqrt{TSS \times HSS} & \quad \mbox{otherwise.}
\end{array}
\right.
\end{equation}

CSS considers both the discrimination power of the model (TSS) and its ability to outperform random chance (HSS), offering a more comprehensive evaluation. It provides a single metric where the values range from 0 to 1, with 1 indicating perfect skill. CSS accounts for both aspects of model performance, making it more suitable for assessing forecast models, particularly in scenarios with class imbalance. Therefore, we evaluate and compare our models based on the single metric, CSS, but we also report both TSS and HSS for completeness.

\subsection{Evaluation}
\begin{figure}[ht!]
\centering
\includegraphics[width=0.95\linewidth ]{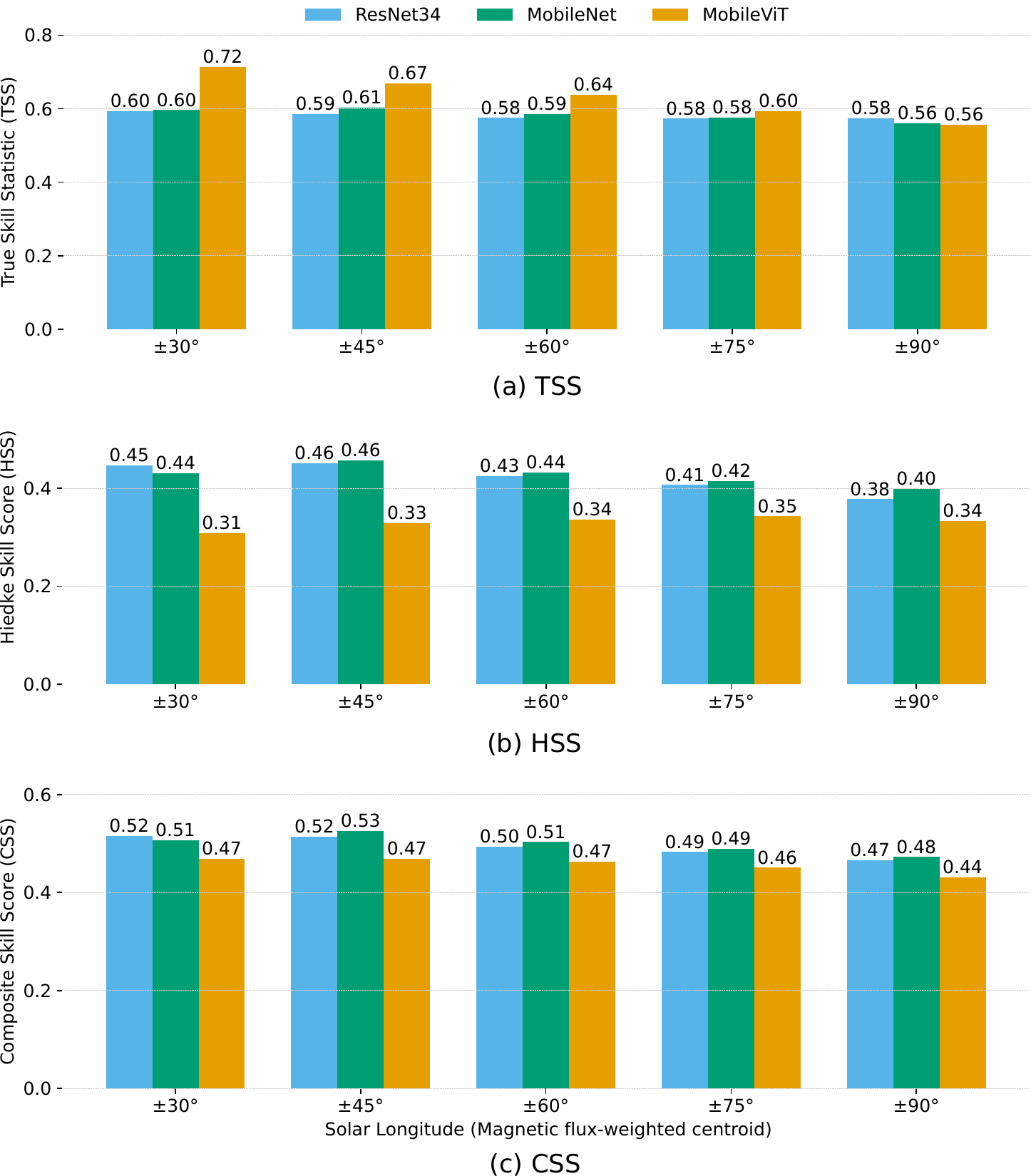}
\caption[]{Performance of our models on test set in terms of (a) TSS, (b) HSS, (c) CSS.}
\label{fig:TSS_HSS_CSS}
% \vspace{-20pt}
\end{figure}
As explained earlier in Sec.~\ref{sec:exptsetup}, we conducted experiments to predict solar flares in a binary setting ($\geq$M-class flares) using a "train-validation-test split" of our entire dataset, which consists of magnetogram of AR patches covering a solar longitudinal range of $\pm$90$^\circ$ (i.e., the entire solar disk). We utilized the validation set to monitor the models' performance every epoch and tuned hyperparameters to optimize the CSS. After training the model with optimal hyperparameters, we employed a threshold tuning approach to calibrate our models by tuning the prediction score thresholds. This involved evaluating the performance of each model on the validation set at different threshold values ranging from 0.01 to 0.99 with an increment of 0.01. We selected the threshold that optimized CSS for each model, resulting in thresholds of 0.36, 0.46, and 0.35 for the ResNet34, MobileNet, and MobileViT models respectively. These thresholds were then applied to the test set for each model, and the models' performance was reported. Additionally, we assessed the performance of all three models on subsets of data representing different longitudinal coverage (within $\pm$30$^\circ$, $\pm$45$^\circ$, $\pm$60$^\circ$, $\pm$75$^\circ$, and $\pm$90$^\circ$ of solar longitude), with $\pm$90$^\circ$ indicating the entire test set. The performance of our models relative to each other in terms of TSS, HSS, and CSS is illustrated in Fig.~\ref{fig:TSS_HSS_CSS} (a), (b), and (c) respectively.

We observed that, the TSS for MobileViT was higher, while HSS was consistently lower compared to other two models, which empirically highlights our hypotheses of using a composite skill score, as choosing the model based on TSS scores might lead to a false sense of good performance. Therefore based on CSS score, we observed that the overall performance corresponding to entire test set ($\pm$90$^\circ$), MobileNet, achieved the best performance with CSS=0.48, (TSS=0.56, HSS=0.40) while the lowest CSS was observed with MobileViT with CSS=0.44 (TSS=0.56, HSS=0.34). Furthermore, our analysis revealed a linearly decreasing trend in model performance with increasing longitudinal coverage of ARs.

\begin{table}[tbh!]
\setlength{\tabcolsep}{6pt}
\renewcommand{\arraystretch}{1.3}
% \vspace{-5pt}
\caption{Comparison with existing literature in terms of TSS, HSS, and CSS with in specific longitudinal coverage.}
% \vspace{-5pt}
\begin{center}
\begin{tabular}{l l c c c }
\hline 

% Longitudinal &  &   &   &\\
Longitudinal Coverage   & Models & TSS  & HSS  & CSS \\
\hline
%  &    &     &    & &    & \\
& Huang et al., 2018 \cite{Huang2018}  & 0.66  & 0.14  & 0.31 \\

& Li et al., 2023 \cite{Li2023} &  0.45 & 0.44  & 0.44\\

Within $\pm$30$^\circ$ & ResNet34 (This Work) & 0.60 & \textbf{0.45} & \textbf{0.52} \\ 
& MobileNet (This Work) & 0.60 & 0.44 & 0.51 \\ 
& MobileViT (This Work) & \textbf{0.72} & 0.31 &0.47 \\ 

\hline
& Bloomfield et al., 2012 \cite{Bloomfield2012} &  0.54 & 0.19  & 0.32\\
& Bobra et al., 2015 \cite{Bobra2015} \textbf{($\pm$68$^\circ$)}&  0.76 & \textbf{0.51}  & \textbf{0.62}\\
& Ji et al., 2022 \cite{Ji2022} \textbf{($\pm$70$^\circ$)}&  \textbf{0.81} & 0.22  & 0.42\\
& Ji et al., 2023 \cite{Ji2023} \textbf{($\pm$70$^\circ$)}&  \textbf{0.81} & 0.43  & 0.59\\
Within $\pm$60$^\circ$ & ResNet34 (This Work) & 0.58 & 0.43 & 0.50 \\ 
& MobileNet (This Work) & 0.59 & 0.44 & 0.51 \\ 
& MobileViT (This Work) & 0.64 & 0.34 & 0.47 \\ 
\hline
& Nishizuka et al., 2018 \cite{Nishizuka2018} &  \textbf{0.80} & 0.26  & 0.45\\
Within $\pm$90$^\circ$ & ResNet34 (This Work) & 0.58 & 0.38 &0.47 \\ 
& MobileNet (This Work) & 0.56 & \textbf{0.40} & \textbf{0.48} \\ 
& MobileViT (This Work) & 0.56 & 0.34 & 0.44 \\ 
\hline
\end{tabular}
\end{center}
\label{table:comp}
% \vspace{-20pt}
\end{table}

We conducted a comprehensive comparison of our models' performance against existing literature within the longitudinal coverage of their experiments as presented in Table.~\ref{table:comp}, focusing on CSS while reporting both the TSS, and HSS metrics as well\footnote{It should be noted that the comparisons against the existing literature might not be accurate due to differences in datasets and partitions used across these studies, as already mentioned in  Sec.~\ref{sec:rel}. However, they are intended to provide general insights into the standings reported in other studies discussed in this paper.}. Notably, the superior scores are highlighted in bold. Our analysis revealed that all three of our models exhibited superior performance compared to both \cite{Huang2018} and \cite{Li2023}. Specifically, ResNet34 surpassed \cite{Huang2018} and \cite{Li2023} by 21\% and 8\% in CSS, respectively, within the $\pm$30$^\circ$ range. Similarly, in comparison to models evaluated within $\pm$60$^\circ$, our model (MobileNet) outperformed \cite{Bloomfield2012} by 19\%. While we show the best performance by CSS in bold for \cite{Bobra2015}, it is important to note the challenges in making precise comparisons due to the varying longitudinal ranges covered in studies by \cite{Bobra2015}, \cite{Ji2022}, \cite{Ji2023} (spanning $\pm$68$^\circ$, $\pm$70$^\circ$, and $\pm$70$^\circ$, respectively). Despite these differences, our findings underscore the effectiveness of our approach within the established longitudinal boundaries.

Furthermore, on evaluating with in $\pm$90$^\circ$, notably, the MobileNet model demonstrated the best results, outperforming \cite{Nishizuka2018} by 3\% in CSS. Finally, we emphasize that although there are full-disk models which covers entire solar disk ($\pm$90$^\circ$), as mentioned in Sec.\ref{sec:rel}, they are not directly comparable to AR- based models \cite{Pandey2022f}, therefore, we exclude full-disk models from comparison. Overall, our study presents a pioneering approach by developing AR-based models that incorporate AR patches spanning the entire solar disk, a methodology never before explored in solar flare prediction. Our comprehensive evaluation demonstrates the superiority of our models over existing ones within their respective longitudinal coverage. This highlights the effectiveness of our novel approach in advancing the solar flare prediction. To reproduce this work, the source code and  experimental results can be accessed from our open source repository \cite{gsudmlab2024arpatchsfp}.

\subsection{Discussion}\label{sec:dis}
  
\begin{figure}[ht!]
% \vspace{-10pt}
\centering
\includegraphics[width=0.95\linewidth ]{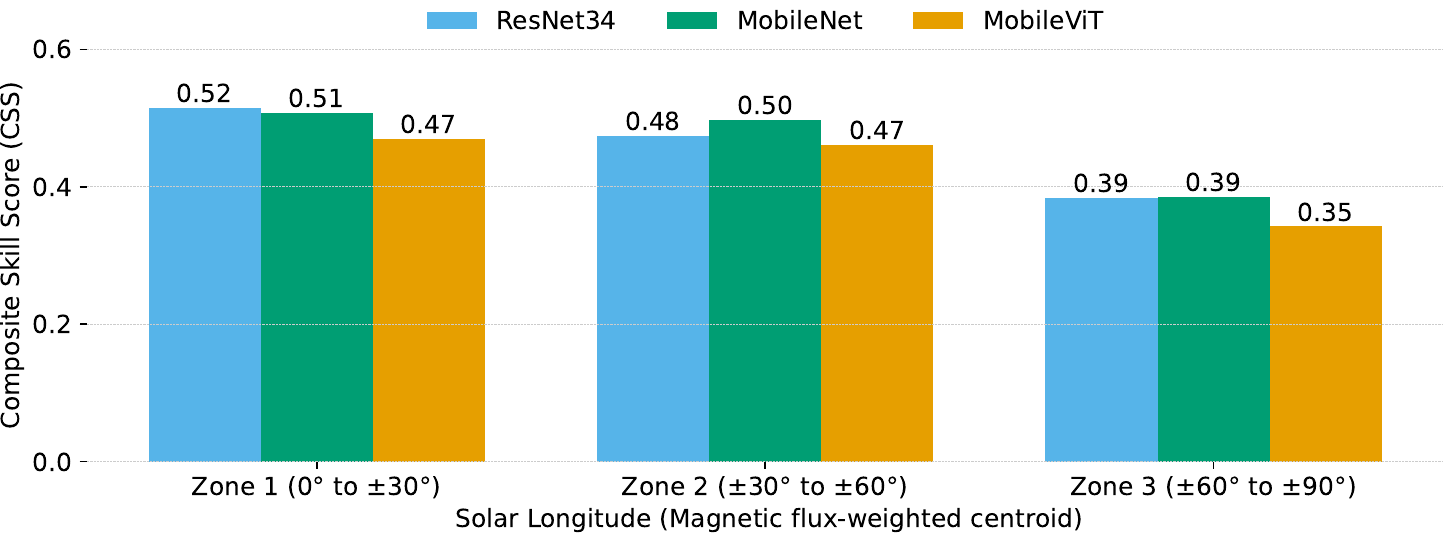}
\caption[]{Models' efficacy evaluated on three different Zones in terms of CSS. Zone 1 and 2 indicates central locations and Zone 3 represents near-limb location.}
\label{fig:zone}
% \vspace{-15pt}
\end{figure}

Upon recognizing a pattern indicating a decline in model performance with increasing longitudinal coverage, we investigated the effectiveness of our models on non-overlapping regions of solar longitudes. To facilitate this analysis, we delineated three zones: Zone 1 representing the region within $\pm$30$^\circ$, Zone 2 covering the area between $\pm$30$^\circ$ to $\pm$60$^\circ$, and Zone 3 spanning $\pm$60$^\circ$ to $\pm$90$^\circ$. To evaluate the models' performance across these zones, we computed the CSS and observed a clear decrease in model skill towards the limb, as illustrated in Fig.~\ref{fig:zone} from all three models. It is worth noting that while existing models are typically designed to predict solar flares up to Zone 2, our model demonstrates capability in the near-limb regions, which includes Zone 3. Despite lower skill scores compared to those in the central region, this study unveils a novel capability that shows skills on the near-limb region, thereby advancing solar flare prediction. This achievement underscores the significance of our research in enhancing our understanding and predictive capabilities in solar phenomena.

\section{Conclusion and Future Work}\label{sec:conc}
In this study, we introduced three limb-to-limb (i.e., $\pm$90$^\circ$) flare prediction models to forecast solar flares of magnitude $\geq$M-class, which are trained with images created from LoS magnetic field component of AR patches. The primary focus of our work was to demonstrate the predictive capacity of AR-based models in near-limb regions (beyond $\pm$60$^\circ$). On evaluating this novel limb-to-limb model capabilities, the results show that we can satisfactorily predict the flaring activity in the existence of severe projection effects although there is room for improvement (the skill is limited when compared to central locations). Furthermore, the results also support our hypothesis that shape-based features derived from magnetograms are effective when predicting solar flares even when the ARs are close to limbs. Moreover, we also introduced a novel preprocessing pipeline for image transformation of magnetic field data products. This pipeline is remarkable for deep learning-based flare forecasting tools and it provides a systematic workflow for clamping, padding, thresholding and most-relevant window selection; which is critical considering the significant class imbalance in forecasting tasks. Finally, we also introduced a new evaluation metric CSS, which is also important and presents us with a practical index for model comparison and selection. While full-disk models are developed to complement AR-based models in near-limb regions, they lack the ability to localize AR-specific predictions. We define this work as a pioneering step towards fully integrating ARs into solar flare prediction, with significant implications for advancing such predictions. Numerous avenues for future exploration exist, including investigating multimodal solar observations, developing spatiotemporal models, and incorporating explanatory/ interpretative frameworks into the model to enhance reliability.\\

\noindent \textbf{Ethical Statement: } Space weather forecasting research involves several ethical considerations. The data for solar flare prediction, provided publicly by NASA/SDO and the AIA, EVE, and HMI science teams, is free from data privacy and security concerns. Non-commercial use of SDO images for education and public information is encouraged without needing authorization. Ethical and responsible development and use of forecasting models are crucial to prevent biases or negative impacts. These models, despite their advanced capabilities, have limitations due to the rarity of extreme solar events and the methods used to evaluate them. Caution and multiple information sources are essential in decision-making related to space weather events. Transparency about these models' limitations ensures ethical and responsible use, mitigating potential harm. Finally, space weather forecasting should be used for peaceful purposes, such as early detection and reducing vulnerabilities to extreme space weather events.

\begin{credits}
\subsubsection{\ackname} 
This work is supported in part under two NSF grants (Award \#2104004 and \#1931555) and a NASA SWR2O2R grant (Award \#80NSSC22K0272). The data used in this study is a courtesy of NASA/SDO and the AIA, EVE, and HMI science teams, and the NOAA National Geophysical Data Center (NGDC).
\end{credits}

\end{document}